\documentclass[aps,pre,superscriptaddress,showkeys,showpacs,twocolumn]{revtex4-1}
\usepackage{graphicx}
\usepackage{latexsym}
\usepackage{amsmath}
\begin {document}
\title {Roulette-wheel selection via stochastic acceptance}
\author{Adam Lipowski}
\affiliation{Faculty of Physics, Adam Mickiewicz University, Pozna\'{n}, Poland}
\author{Dorota Lipowska}
\affiliation{Faculty of Modern Languages and Literature, Adam Mickiewicz University, Pozna\'{n}, Poland}
%%%%%%%%%%%%%%%%%%%%%%%%%%%%%%%%%%%%%%%%%%%%%%%%%%%%%%%%%%%%%%%%%%%%%%%%%%%%%
\begin {abstract}
Roulette-wheel selection is a frequently used method in genetic and evolutionary algorithms or in modeling of complex networks. Existing routines select one of   $N$ individuals using search algorithms of $O(N)$ or $O(\log N)$ complexity. We present a simple roulette-wheel selection algorithm, which typically has $O(1)$ complexity and is based on stochastic acceptance instead of searching. We also discuss a hybrid version, which might be suitable for highly heterogeneous weight distributions, found, for example, in some models of complex networks. With minor modifications, the algorithm  might also be used for sampling with  fitness cut-off at a certain value or for sampling without replacement. 
\end{abstract}

\maketitle
\section{Introduction}
Finding low-energy configurations of crystals~\cite{crystal}, disordered magnets~\cite{spinglass} or proteins~\cite{proteins}, reconstructing geological structure from seismic data \cite{seismic}, and analysing X-ray data \cite{xray} are just a few examples of optimization problems in physical sciences \cite{rieger}. Various techniques have been developed to approach these problems and one of the most frequently used is genetic algorithms \cite{holland}. Their basic idea is to mimic the way biological evolution creates apparrently better fitted species. To do so one needs to represent a pool of optimization methods (routines, functions, strategies, etc.) as a population of individuals, which, as in nature, is subjected to two, in a sense opposing, processes. On the one hand, due to mutations or crossing-over operations, the variability in the population increases. On the other hand, to guide the evolution in a desired direction, one has to trim the population with some selection mechanisms. It turns out that such a biology-inspired scheme allows us to find optimal or nearly optimal solutions of  various problems in a very efficient way~\cite{goldberg}.

\begin{figure}
\vspace{0.7cm} 
\includegraphics[width=9cm]{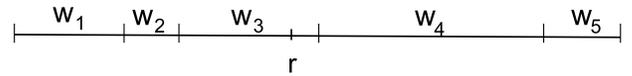}  \vspace{-0.7cm} 
\caption{ 
In a roulette-wheel selection, one constructs  a line segment of length $\sum_{i=1}^N w_i$ out of consecutive sectors of length $w_i\ (i=1,2,\ldots,N)$, generates a random number $r$ such that $0<r<\sum_{i=1}^N w_i$, and locates the corresponding sector ($w_3$ in this case), thus selecting the respective individual. When a simple search of an $N$-element list is used for location, the procedure requires $O(N)$ operations, which can be reduced to $O(\log N)$ by means of the binary search.
\label{ruletka}}
\end{figure}

Selection is an important part of genetic algorithms since it  affects  significantly their convergence. The basic strategy follows the rule: The better fitted an individual, the larger the probability of its survival and mating. The most straightforward implementation of this rule is the so-called roulette-wheel selection \cite{goldberg}. This method assumes that the probability of selection is proportional to the fitness of an individual. It can be briefly described as follows. Let us consider $N$ individuals,  each characterized by its fitness $w_i>0$\ $(i=1,2,\ldots,N)$. The selection probability of the $i$-th individual is thus given as 
\begin{equation}
p_i=\frac{w_i}{\sum_{i=1}^N w_i} \ \ \ \ \ \ \ \ (i=1,2,\ldots,N)
\label{prob}
\end{equation}
Let us imagine a roulette wheel with sectors of size proportional to  $w_i \ (i=1,2,\ldots,N)$. Selection of an individual  is then equivalent to choosing randomly a point on the wheel and locating the corresponding sector (Fig.~\ref{ruletka}).
When a simple search is used, such a location requires $O(N)$ operations while the binary search needs $O(\log N)$ operations.

%%%%%%%%%%%%%%%%%%%%%%%%%%%%%%%%
%%%%%%%%%%%%%%%%%%%%%%%%%%%%%

There are also other methods with the selection probability depending on the fitness, such as stochastic remainder or stochastic universal selection \cite{baker87,goldberg-deb}. They have slightly different statistical properties and in general lead to populations of larger variability. In yet another class of selection methods, the ranking or ordering of individuals rather than their fitness plays a central role~\cite{baker}.  It is certainly difficult to indicate which of these methods is the best~\cite{sivaraj,razali}. Although some studies suggest that fitness-based selection methods suffer from certain scaling-related problems \cite{hancock},  there are some works that seem to alleviate this difficulty \cite{gupta}. 

Let us notice that the roulette-wheel selection is also used in the generation of complex networks.  For example, in some models of growing networks, a newly added site is linked to one of the already existing sites with probability proportional to the degree of this old site \cite{barab,krap,mendes}. This preferential-attachment algorithm is known to generate highly heterogeneous scale-free networks, which recently have been intensively studied~\cite{boccaletti}. The roulette-wheel selection is also used in certain adaptive network models~\cite{liplip}.

Due to simplicity of implementation and straightforward interpretation, the roulette-wheel selection is frequently used in genetic algorithms. Although the binary search significantly reduces computational complexity, still faster implementations would be, in our opinion, desirable. In the present paper, we show that the roulette-wheel selection can be realized with a simple algorithm of typically $O(1)$ complexity. The proposed algorithm does not use searching but is based on a stochastic acceptance of a randomly selected individual.
%%%%%%%%%%%%%%%%%%%%%%%%%%%%%%%%
\section{Description and properties of the algorithm}
\begin{figure}
\includegraphics[width=9cm]{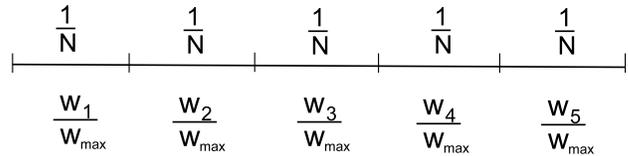}  \vspace{-0.7cm} 
\caption{ 
In the proposed algorithm, one selects randomly (with equal probability $1/N$) an individual (say, $i$) and accepts such a selection with probability $w_i/w_{\rm max}$, where $w_{\rm max}$ is the maximal fitness. In case of rejection, the procedure is repeated anew (i.e., a new individual is selected repeatedly until acceptance). Although it requires extra call(s) of a random-number generator, it remains  typically $O(1)$ algorithm.
\label{ruletka-new}}
\end{figure}

Our algorithm consists of the following steps:
\begin{enumerate}
\item Select randomly one of the individuals (say, $i$). The selection is done with uniform probability ($1/N$), which does not depend on the  individual's fitness $w_i$  (Fig.~\ref{ruletka-new}).
\item With probability $w_i/w_{\rm max}$, where $w_{\rm max}={\rm max}\{w_i \}_{i=1}^N $ is the maximal fitness in the population, the selection is accepted. Otherwise, the procedure is repeated from step~1 (i.e., in the case of rejection, another selection attempt is made).
\end{enumerate}

Of course, the probability that the $i$-th individual will be selected in a single attempt equals $w_i/(Nw_{\rm max})$. However, since several first attempts might fail, one has to calculate the probability that the  individual selected in an unspecified number of attempts will be the $i$-th individual. Straightforward calculations give
\begin{equation}
p_i'=\frac{w_i}{Nw_{\rm max}}(1+q+q^2+\ldots)
\label{series}
\end{equation} 
where
\begin{equation} 
q=\frac{1}{N}\sum_{i=1}^N (1-w_i/w_{\rm max})=1-\frac{\sum_{i=1}^N w_i}{Nw_{\rm max}}=1-\frac{\langle w \rangle}{w_{\rm max}}
\end{equation}
is the rejection probability and $\langle w \rangle=(\sum_{i=1}^N w_i)/N$ is the average fitness. The geometrical series (\ref{series}) is convergent ($0<q<1$) and summing it up, we easily find that 
\begin{equation}
p_i'=p_i
\label{pp}
\end{equation}
where $p_i$ is defined in Eq.~(\ref{prob}). This shows that the probability distribution of our procedure is indeed the same as in the roulette-wheel selection.

Similarly, we can calculate the average number of attempts $\tau$ needed for a single selection of any individual. One obtains
\begin{equation}
\tau=\frac{1}{N}\sum_{i=1}^N \frac{w_i}{w_{\rm max}} [1+2q+3q^2+\ldots]=\frac{w_{\rm max}}{\langle w \rangle}
\label{tau}
\end{equation}

Although $\tau$, which determines the computational complexity of our algoritmm, does not depend explicitly on $N$, the ratio $\frac{w_{\rm max}}{\langle w \rangle}$ might change  with $N$, depending on the specificity of the problem. We expect, however, that in many applications, for example, those where fitness remains bounded ($w_i<B$) and $\langle w \rangle$ does not diminish to 0 for increasing $N$, the ratio $\frac{w_{\rm max}}{\langle w \rangle}$ does not increase unboundedly with $N$ and thus a typical complexity of our algorithm should be~$O(1)$.

One can examine a more general algorithm where acceptance is made with probability $w_i/A$, where $A>w_{\rm max}$ is a certain constant. For such an algorithm the selection probability $p_i'$ also satisfies Eq.~(\ref{pp}). However, a smaller  acceptance probability  results in a larger rejection probability and the efficiency of the algorithm is reduced (as $\tau$ increases). For $A<w_{\rm max}$, some of the fractions $w_i/A$ may  turn out  greater than unity and the equality~(\ref{pp}) does not hold. Thus, the choice $A=w_{\rm max}$ ensures  optimal performance.

%%%%%%%%%%%%%%%%%%%%%%%%%%%%%%%%
%%%%%%%%%%%%%%%%%%%%%%%%%%%%%

\begin{figure}
\vspace{0.5cm} 
\includegraphics[width=9cm]{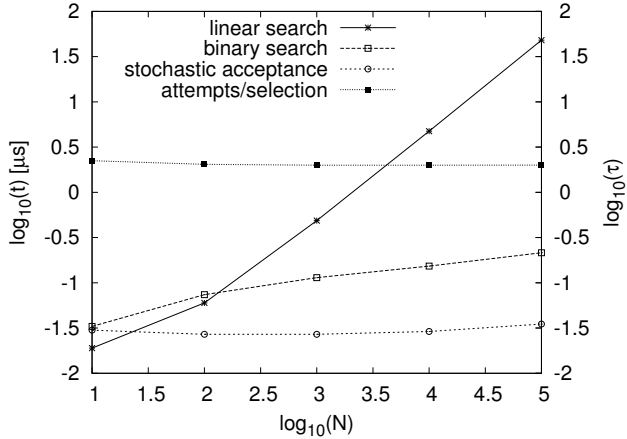}  \vspace{-0.7cm} 
\caption{ 
The average CPU time $t$ (in $\mu$s) of roulette-wheel selection of a single individual as a function of the population size~$N$ (log-log scale). As expected, the linear search has the complexity $O(N)$ and the binary search $O(\log(N))$. The stochastic acceptance algorithm behaves as $O(1)$ and a slight increase for the  largest~$N$ is due to excessive memory requirements that exceeded the size of cache memory. Calculations were made on the Intel Xeon 2.8GHz processor. Filled squares show the average number of attempts per selection.
\label{avtime}}
\end{figure}

To examine its performance, we applied our algorithm to a population of $N$ individuals with fitness being a uniformly distributed random number   $0<w_i<1$. We implemented the roulette-wheel selection algorithms using linear or binary search as well as our stochastic acceptance method. We found that the distributions of selected individuals generated by these programs were within statistical errors the same. Average execution time per single selection as a function of $N$ is shown in Fig.~\ref{avtime}. As expected, the linear and binary search show $O(N)$ and $O(\log N)$ behaviour, respectively. Our algorithm requires basically $N$-independent CPU time and a slight increase for large $N$ is due to excessive memory requirements that  exceeded the size of cache memory.
We also measured the average number of attempts $\tau$. In our numerical example, $w_{\rm max}\approx 1$ for large~$N$, and $\langle w \rangle \sim 1/2$. Thus, according to Eq.~(\ref{tau}), $\tau\approx 2$, and our numerical data are in excellent agreement with this result (Fig.~\ref{avtime}).
%%%%%%%%%%%%%%%%%%%%%%%%%%%%%%%%
%%%%%%%%%%%%%%%%%%%%%%%%%%%%%
%%%%%%%%%%%%%%%%%%%%%%%%%%%%%
\section{Possible extensions and conclusions}
As we have already mentioned in Introduction, the roulette-wheel selection is used in some models of complex networks that are based on preferential attachment. The  networks generated in this way are strongly inhomogeneous with a small fraction  of sites having a very large degree. The distribution of weights, which are proportional to degrees, is thus very broad and the  ratio $\frac{w_{\rm max}}{\langle w \rangle}$ might be quite large. Consequently, the efficiency of our algorithm might diminish. However, a simple modification of the  algorithm might lead to a better performance. First, let us assume that there is one weight that is much larger than the others, say $w_1$ ($=w_{\rm max}\gg w_i,\ \  i=2,3,\ldots N$). 
In this case we might use the following hybrid algorithm, which combines search and stochastic acceptance: With probability $p_1=w_1/\sum_{i=1}^N w_i$ one selects the first individual and with probability $1-p_1$ one of the remaining $N-1$ individuals (using roulette-wheel applied to $N-1$ weights). In the latter step, the  stochastic  acceptance should be quite efficient since the largest weight $w_1$ was removed. 
The probability of selection of the individual $i(>1)$ equals $(1-p_1)\frac{w_i}{\sum_{j=2}^N w_j}=\frac{w_i}{\sum_{j=1}^N w_j}$, thus it is indeed equal to~(\ref{prob}).
Generalization to the case where there are several much larger weights  is straightforward.

To ensure a sufficiently large variability of the population, it is sometimes desirable in optimization problems to  use sampling without replacement. In order to guarantee that a once selected individual is never selected again, one can simply set its fitness to 0. However, when the individual happened to have the maximal fitness, a new maximum should be found. Sacrifying slightly the efficiency, one can use the old maximum to calculate the acceptance probability $w_i/w_{\rm max}$. Similarly, when fitness of individuals is known to be bounded ($w_i<B$), it is not needed to keep track of the current maximum, as one can use $w_i/B$ as the acceptance probability. Although the efficiency of the algorithm is reduced, it still should remain of the $O(1)$ type \cite{comment1}. 

The replacement of $w_{\rm max}$ with a certain constant $A<w_{\rm max}$  in our algorithm might be yet another way of increasing variability in the  population. Indeed, in such a case individuals are selected according to Eq.~(\ref{prob}) but with their fitness cut-off at $A$ (i.e., $w_i'= {\rm min}\{w_i,A\}$). Our algorithm might also  be adapted to evolving systems, such as, for example, complex adaptive systems where fitness of some or all individuals changes in time. Let us emphasize that the performance of a selection method depends on a number of factors and also on a particular type of the optimization problem \cite{razali}. More detailed analysis of the  efficiency of our algorithm in comparison with other selection methods will be presented elsewhere.

In conclusion, we have shown that Holland's original idea of using fitness-proportionate selection, i.e., the so-called roulette-wheel selection, can be formulated as an algorithm of typically $O(1)$ complexity. The numerical example shows that for sizes of populations used in genetic-algorithm applications, ranging from $10^2$ to $10^4$, our algorithm offers a  significant CPU gain over exisiting routines based on a linear or binary search. The algorithm is very simple and we expect that it can be modified and used in even more sophisticated selection schemes. 

%Let us notice that for any $i$ and $j$ we have $p_i'/p_j'=p_i/p_j$, where $p_i$ and $p_j$ %satisfy (\ref{prob}).

%%%%%%%%%%%%%%%%%%%%%%%%%%%%%%%%%%%%%%%%%%%%%%%%%%%%%%%%%%%

%%%%%%%%%%%%%%%%%%%
%%%%%%%%%%%%%%%%%%%%%%%%%%%%%%%%%%%%%%%%%%%%%%%%%%%%%%%%%%%%%%%%%%%%%%%%%%%%%%%
%%%%%%%%%%%%%%%%%%%%%%%%%%%%%%%%%%%%%%%%%%%%%%%%%%%%%%%%%%%%%%%%%%%%%%%%%%%%%%%

\begin{thebibliography}{}
\bibitem{comment1} Genetic algoritms are unlikely to operate in a regime of vanishing $\langle w\rangle$ and thus some  bounds imposed on fitness should result in finite $\tau$ and $O(1)$ complexity of the algorithm.
\end{thebibliography}
\end {document}